\documentclass[3p]{elsarticle}

\usepackage{lineno,hyperref}
\usepackage{multirow}
\usepackage{amsfonts}
\usepackage{bm}
\usepackage{subfigure}
\usepackage{booktabs}
\usepackage{amsmath}
\usepackage{graphicx}

\modulolinenumbers[5]

\journal{Neurocomputing}

\begin{document}

\begin{frontmatter}

\title{Efficient Spatiotemporal Context Modeling for Action Recognition\tnoteref{titlenote}}
\tnotetext[titlenote]{Funding: This work was supported by the National Natural Science Foundation of China [grant numbers 61906155, U19B2037]; the Natural Science Foundation of Shaanxi Province [grant number 2020JQ-216]; and the China Postdoctoral Science Foundation [grant number 2020M673488].}


%
%

\author[1]{Congqi~Cao\corref{cor1}\fnref{fn1}}
\ead{congqi.cao@nwpu.edu.cn}

\author[1]{Yue~Lu\fnref{fn1}}
\ead{zugexiaodui@mail.nwpu.edu.cn}

\author[2]{Yifan~Zhang\fnref{fn2}}
\ead{yfzhang@nlpr.ia.ac.cn}

\author[1]{Dongmei~Jiang\fnref{fn1}}
\ead{jiangdm@mail.nwpu.edu.cn}

\author[1]{Yanning~Zhang\fnref{fn1}}
\ead{ynzhang@nwpu.edu.cn}

\cortext[cor1]{Corresponding author}

\fntext[fn1]{National Engineering Laboratory for Integrated Aero-Space-Ground-Ocean Big Data Application Technology and the Shaanxi Provincial Key Lab on Speech and Image Information Processing, School of Computer Science, Northwestern Polytechnical University, Xi'an 710129, China}

\fntext[fn2]{National Laboratory of Pattern Recognition, Institute of Automation, Chinese Academy of Sciences, and the University of Chinese Academy of Sciences, Beijing 100190, China}

\address[1]{School of Computer Science, Northwestern Polytechnical University, Xi'an, China}
\address[2]{National Laboratory of Pattern Recognition, Institute of Automation, Chinese Academy of Sciences, Beijing, China}

\begin{abstract}
Contextual information plays an important role in action recognition. Local operations have difficulty to model the relation between two elements with a long-distance interval. However, directly modeling the contextual information between any two points brings huge cost in computation and memory, especially for action recognition, where there is an additional temporal dimension. Inspired from 2D criss-cross attention used in segmentation task, we propose a recurrent 3D criss-cross attention (RCCA-3D) module to model the dense long-range spatiotemporal contextual information in video for action recognition. The global context is factorized into sparse relation maps. We model the relationship between points in the same line along the direction of horizon, vertical and depth at each time, which forms a 3D criss-cross structure, and duplicate the same operation with recurrent mechanism to transmit the relation between points in a line to a plane finally to the whole spatiotemporal space. Compared with the non-local method, the proposed RCCA-3D module reduces the number of parameters and FLOPs by 25\% and 30\% for video context modeling. We evaluate the performance of RCCA-3D with two latest action recognition networks on {three} datasets and make a thorough analysis of the architecture, obtaining the optimal way to factorize and fuse the relation maps. Comparisons with other state-of-the-art methods demonstrate the effectiveness and efficiency of our model.
\end{abstract}


\end{frontmatter}


\section{Introduction}

Action recognition, which aims to assign action labels to each video sequence, has a wide application in human-computer interaction, motion analysis and synthesis, intelligent monitoring, content-based video retrieval and so on.
The application of deep neural networks \cite{krizhevsky2012imagenet, He2016CVPR} has achieved great success in the field of action recognition \cite{tran2015learning, Carreira2017CVPR, wang2018temporal, xie2018rethinking}.
Convolutional Neural Networks (CNNs) are extensively employed to extract visual features. {Since the basic 2D CNNs, which extract appearance features of each frame separately, cannot capture the motion information in the video,} two-stream CNNs \cite{simonyan2014two, wang2018temporal} are proposed to use the stacked optical flow images as an additional input for short-term motion modeling. In order to model the long-term motion information in video, Recurrent Neural Networks (RNNs) are combined with 2D CNNs for video sequence modeling \cite{Donahue2015CVPR}.
Different from 2D CNNs, 3D CNNs \cite{tran2015learning} capture spatial and temporal information at the same time with 3D spatiotemporal kernels, which are more suitable for video analysis. They are widely used in the field of action recognition, and many improved methods based on 3D CNNs are derived \cite{Carreira2017CVPR, xie2018rethinking, tran2018closer}.
However, 3D CNNs have the problem of high computational complexity. Some methods aim at reducing the computational complexity of 3D CNNs. Separable 3D CNN \cite{xie2018rethinking} factorizes 3D convolution into 2D and 1D convolutions. Multi-Fiber Network \cite{Chen2018ECCV} reduces considerable computation by using improved 3D group convolution.
Temporal Shift Module \cite{lin2019tsm} shifts features along time, and then uses 2D CNNs to process the shifted feature maps at each time.
The above works are based on local operations for information extraction and short-term contextual modeling, which expand the receptive field by stacking multiple layers.

\begin{figure}[tp]
	\centering
	\includegraphics[width=0.7\columnwidth]{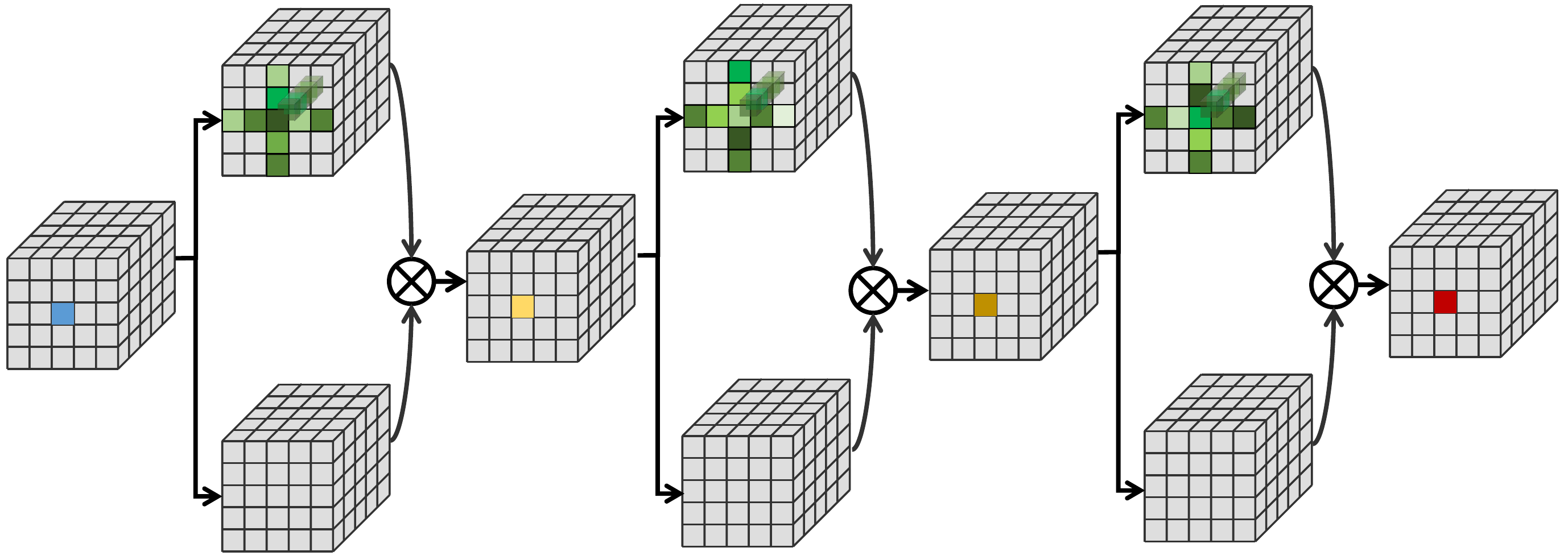}
	\caption{Illustration of our recurrent 3D criss-cross attention module. For each position (e.g. blue), a 3D criss-cross attention module generates a sparse attention map. After a series of recurrent operations, a 3D feature map with dense contextual information from all positions is obtained.
	}
	\label{fig1}
\end{figure}

In the real world, the object does not exist alone, it always has relationships with other objects or the environment. In order to recognize an action, we need to combine the visual features of the object and its interactions with other objects or the scene. The interactions may occur between distant pixels in space as well as time \cite{wang2018non}. Therefore, modeling the long-range dense context is essential.
The most common method for long-range and dense contextual information modeling is to repeat the local operations, \textit{i.e.} the convolutional and recurrent operations. However, repeating local operations is inefficient and causes optimization difficulties.

In order to enhance the ability of capturing long-term contextual dependencies, non-local neural network is proposed \cite{wang2018non}. A non-local module directly computes the correlations between every two points in a spatiotemporal feature map by matrix multiplication. Several non-local modules are plugged into the deep 3D convolutional neural networks to capture dense long-range dependencies and thus improve the recognition accuracy. However, non-local operation causes a relative high computation cost.
For a clip of video with $\mathit{T}$ frames, the computation complexity of the non-local module is $\mathcal{O}$$\mathit{((T}$$\mathit{\times}$$\mathit{H}$$\mathit{\times}$$\mathit{W}$$\mathit{\times}$$\mathit{C)}$$\mathit{\times(T}$$\mathit{\times}$$\mathit{H}$$\mathit{\times}$$\mathit{W}$$\mathit{\times}$$\mathit{C))}$, where $\mathit{H}$, $\mathit{W}$ and $\mathit{C}$ represent the height, width and the number of channels of the feature map respectively.
Double attention network (\textit{abbr}. $A^2$-Net) \cite{NIPS20187318} aggregates global features into a compact set and redistributes features via two attention processes, which can be seen as another implementation of non-local operation with fewer computation. The computation complexity of $A^2$-Net is $\mathcal{O}$$\mathit{((T}$$\mathit{\times}$$\mathit{H}$$\mathit{\times}$$\mathit{W}$$\mathit{\times C)}$$\mathit{\times C)}$, but usually $\mathit{C}$ is very large.
Compact generalized non-local (CGNL) network \cite{yue2018compact} approximates non-local operation via low-order Taylor series. Its computation complexity is $\mathcal{O}$$\mathit{((T}$$\mathit{\times}$$\mathit{H}$$\mathit{\times}$$\mathit{W}$$\mathit{\times C)}$$\mathit{\times (P+}$$\mathrm{1}$$\mathit{))}$, where $\mathit{P}\ll\mathit{(T}$ $\mathit{\times}$$\mathit{H}$ $\mathit{\times}$$\mathit{W})$ indicates the order of Taylor expansion for kernel functions.
In the field of image segmentation, criss-cross attention (CCA) \cite{Huang_2019_ICCV} is proposed to reduce the computation complexity of non-local operation. One CCA module harvests sparse and pixel-wise contextual information on the criss-cross path for all the points in a feature map. With two stacked CCA modules in series, every point in the feature map is able to interact with any other point, which generates dense contextual information and long-range dependencies of the whole feature map. For a 2D feature map, the computation complexity of non-local operation is $\mathcal{O}$$\mathit{((H}$$\mathit{\times}$$\mathit{W}$$\mathit{\times}$$\mathit{C)}$$\mathit{\times(H}$$\mathit{\times}$$\mathit{W}$$\mathit{\times C))}$, while that of the CCA module is $\mathcal{O}$$\mathit{((H}$$\mathit{\times}$$\mathit{W}$$\mathit{\times}$$\mathit{C)}$$\mathit{\times(H+W-}$ $\mathrm{1}$$\mathit{))}$ \cite{Huang_2019_ICCV}. Criss-cross attention has shown an efficient and strong ability of spatial context modeling in image segmentation task.

In our work, we build 3D criss-cross attention modules and insert them into multiple state-of-the-art action recognition architectures.
For a spatiotemporal feature map, a 3D criss-cross operation only needs $(T+H+W-2)$ sparse connections for each point, which reduces vast computation complexity and memory usage. To make the interaction between any two points in the spatiotemporal domain possible, a recurrent 3D CCA (RCCA-3D) module that includes three CCA-3D modules with shared parameters and recurrent connections is proposed, which can be seen in Figure \ref{fig1}. The complexity of RCCA-3D module is $\mathcal{O}$$\mathit{((T}$$\mathit{\times}$$\mathit{H}$$\mathit{\times}$$\mathit{W}$$\mathit{\times}$$\mathit{C)}$$\mathit{\times(T+H+W-}$ $\mathrm{2}$$\mathit{))}$.
To find the optimal and efficient structure for video context modeling, we conduct an exhaustive research on different network structures for the action recognition tasks. First, we compare different combinations of the CCA-3D modules and determine the most effective structure of the RCCA-3D module. We make a series of explorations on the structure of the RCCA-3D as well as the fusion method between the relation representation obtained by our CCA-3D module and the appearance representation of the input to find the most effective architecture of the RCCA-3D module.
Then we explore the inserted position of the RCCA-3D and the number of inner channels for each CCA-3D module, so as to help us make a trade-off between the accuracy and the amount of calculation and parameters.
Finally, we compare our RCCA-3D module with the non-local module in neural networks of different architectures and on different datasets to verify the superiority of our method. Compared with the non-local method, the proposed RCCA-3D can reduce the number of parameters by 25\% and reduce the computation amount FLOPs by 30\%. With substantial experiments on the UCF101 \cite{soomro2012ucf101}, HMDB51 \cite{kuehne2011hmdb},
Mini-Kinetics and Kinetics-400 \cite{Carreira2017CVPR} datasets, we prove that our proposed RCCA-3D module is effective in spatiotemporal context modeling.

The main contributions of our work include:
\begin{itemize}
	\item
	We extend 2D criss-cross attention to 3D. A CCA-3D module can model the relationship between the spatiotemporal points in the same line along the direction of horizon, vertical and depth at each time with a sparse attention map.	
	\item
	We explore the connections between multiple CCA-3Ds, number of CCA-3Ds and other settings. We design a 3D recurrent criss-cross attention module that is most suitable to model the long-range spatiotemporal contextual information for action recognition.
	\item
	Extensive experiments with different backbones on multiple datasets are made to thoroughly analyze relation models and validate the efficiency and effectiveness of our proposed model, which achieves leading performance among the methods that only use RGB modality.
\end{itemize}

\section{Related Work}

\subsection{Action Recognition}
The spatiotemporal information is the key to video-based action recognition task. To extract short-term and long-term temporal features as well as spatial features, combining optical flow and recurrent operation with 2D convolutional operation are two normal strategies.
Two-stream CNNs \cite{simonyan2014two, Carreira2017CVPR, wang2018temporal} combine a temporal stream that uses optical flow images as input with a spatial stream that uses RGB frame as input for spatial and temporal information fusion.
The long-term recurrent convolutional network \cite{Donahue2015CVPR} (LRCN) uses long short-term memory \cite{Graves1997Long} (LSTM) layers after convolutional layers to model the temporal information. However, such a combination of CNN and LSTM brings huge computation and the performance of the model is far from satisfactory.
Feature Aggregation for SpatioTEmporal Redundancy \cite{zhu2019faster} (FASTER) can integrate subtle motion information from expensive models with scene changes from cheap models and aggregate different representations. Although FASTER utilizes recurrent operations to model temporal relations, it is more lightweight than LRCN.

Another way of extracting spatiotemporal features is to employ 3D convolutional neural networks. C3D \cite{tran2015learning} extends the 2D convolutional kernels in VGG \cite{simonyan2014very} to 3D, but this causes a heavy calculation burden and a mass of parameters, which is highly likely to engender over-fitting. Inflated 3D CNN \cite{Carreira2017CVPR} (I3D) achieves higher accuracy in video classification using multi-scale 3D convolutional kernels, which further increases computational cost and memory consumption. Separable 3D CNN \cite{xie2018rethinking} (S3D) reduces calculation through separating 3D convolutional kernels. Pseudo-3D Residual Net \cite{qiu2017learning} (P3D ResNet) divides spatiotemporal convolutional kernels into spatial and temporal domains, and exploits different combinations of the two domain blocks.
Multi-fiber network \cite{Chen2018ECCV} (MF-Net) employs group convolution with sparse connections into 3D residual blocks to decrease the amount of parameters and computation cost of 3D CNNs. Besides, it develops a multiplexer to compensate the information loss.
Temporal Shift Module (TSM) \cite{lin2019tsm} replaces the 3D convolution with temporal shift operation and 2D convolutions. It first stacks the frames as a clip of video, and then shift a part of frames along the temporal dimension. Finally it splits the clip into frames. Since TSM is implemented by 2D convolution network and the shift operation which does not give rise to additional computation, it alleviates considerable computational burden.

We choose MF-Net and TSM as the backbones, because they have enough ability of feature extraction in action recognition task while maintain lightweight. In addition, they are both convenient to insert extra self-contained modules.

\begin{figure*}[tp]
	\centering
	\includegraphics[width=\textwidth]{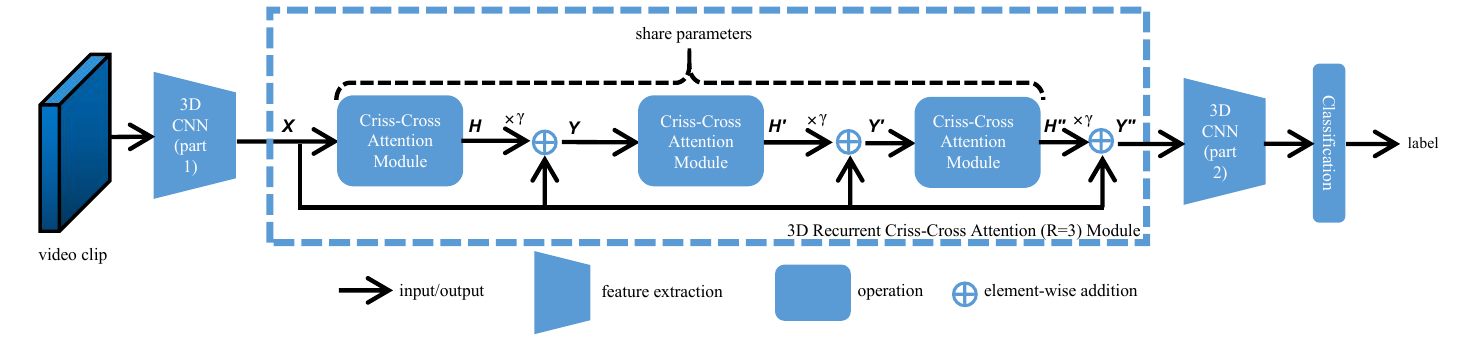}
	\caption{Overview of the network with RCCA-3D module. The RCCA-3D module includes three CCA-3D modules in series. Element-wise addition is applied for the output of each CCA-3D module and the input of the RCCA-3D module. Every CCA-3D module collects sparse contextual information and shares parameters. As a whole, the recurrent criss-cross attention module is able to generate new feature maps with rich and dense contextual information from all positions.
	}
	\label{fig2}
\end{figure*}

\subsection{Relation Model}
Long-range contextual information plays an important role in intelligent analysis. However, it is difficult to capture and model that by simply stacking multiple local operation-based layers for complex scenarios.

Temporal Relation Network \cite{zhou2018temporal} (TRN), which is inspired by the relational network \cite{santoro2017simple}, learns temporal relations at multiple time scales.
Spatio-Temporal Context Model \cite{stcm} (STCM) simultaneously models the before-action, after-action and action itself to integrate temporal contextual information. It also collects spatial contextual information by modeling the whole frame and capturing the relations between the target human action region and its surrounding region.
Based on ConvLSTM \cite{convlstm}, Correlational Convolutional LSTM \cite{cclstm} (C\textsuperscript{2}LSTM) computes the correlation between the input matrices of t-1 and t, by dividing them into patches and computing the cross correlation of each pair.
Convolutional relation network \cite{convrn} stacks convolutional relation blocks, which includes dilation group-specific convolution, temporal convolution and attention pooling, to extract global features of node pairs and edge pairs for skeleton-based action recognition.
\cite{bilstmcnn} uses the relative position and relative speed to provide spatial information and dynamic information in skeletal action sequences, and build a bidirectional LSTM-CNN to fuse spatial–temporal information for skeleton-based action recognition.
Spatial-Temporal Pyramid Network \cite{stpnet} (S-TPNet) extracts multi-scale appearance features from different stages of the 2D CNN. It groups the frame features into different snippet-level features and uses fully-connected layers to reason about snippet relations.

Another efficient way to model long-range relations is the self-attention mechanism.
The self-attention mechanism used to model long-range relations between any two elements is first applied in machine translation \cite{vaswani2017attention}.
A self-attention module computes the response at a position in a sequence by attending to all positions and taking their weighted average in an embedding space.
Non-local neural network \cite{wang2018non} associates self-attention for machine translation with the more general class of non-local filtering operations that are applicable to computer vision.
Double attention network \cite{NIPS20187318} captures the second-order feature statistics and makes adaptive feature assignment by gathering and distributing long-range features.
Compact generalized non-local network (CGNL) \cite{yue2018compact} proposes the mechanism to model the interactions between positions across channels in non-local modules, making it in a fast and low-complexity computation flow with low-order Taylor series approximation.
Expectation-maximization attention network \cite{Li_2019_ICCV} formulates the attention mechanism into an expectation-maximization manner, which {helps it to be friendly} in memory and computation and be robust to the variance of input.
Global Context Network \cite{cao2019gcnet} (GCNet) uses a simplified query-independent non-local network to reduce computation cost, and it unifies 
Squeeze-Excitation Network \cite{hu2018squeeze} (SENet) and non-local network into a general framework for global context modeling.
Video Action Transformer Network \cite{girdhar2019video} uses Transformer-style \cite{vaswani2017attention} architecture to attend relevant regions of the person to their context.
Criss-cross attention module \cite{Huang_2019_ICCV} harvests the contextual information of its surrounding pixels on the criss-cross path to realize the self-attention mechanism with less computation, and performs better in image segmentation task than non-local networks. However, it has not been applied to and verified in action recognition task.
We extend the CCA to CCA-3D and design a novel architecture of the RCCA-3D. Each CCA-3D module can capture the long-distance spatial and temporal dependencies at the same time, so as to improve the accuracy for action recognition task.

\section{Method}

The overview of our proposed method is shown in Figure \ref{fig2}. The input is a clip of video with $T$ frames. After the video clip is fed into the first part of a 3D convolutional network, a 3D feature map \textbf{\textit{X}} with size $(C, T, H, W)$ is generated, where $H, W$ represent the height and width respectively, and $C$ represents the number of channels. In order to obtain long-term dependencies between all the pixels, we feed \textbf{\textit{X}} into the proposed 3D recurrent criss-cross attention module (RCCA-3D) for context modeling. An RCCA-3D module is composed of three CCA-3D modules. Each CCA-3D module only needs to capture sparse 3D criss-cross attention information. By stacking multiple CCA-3D modules, the RCCA-3D module gets dense long-term spatiotemporal relation information.

\begin{figure*}[t]
	\centering
	\includegraphics[width=0.6\textwidth]{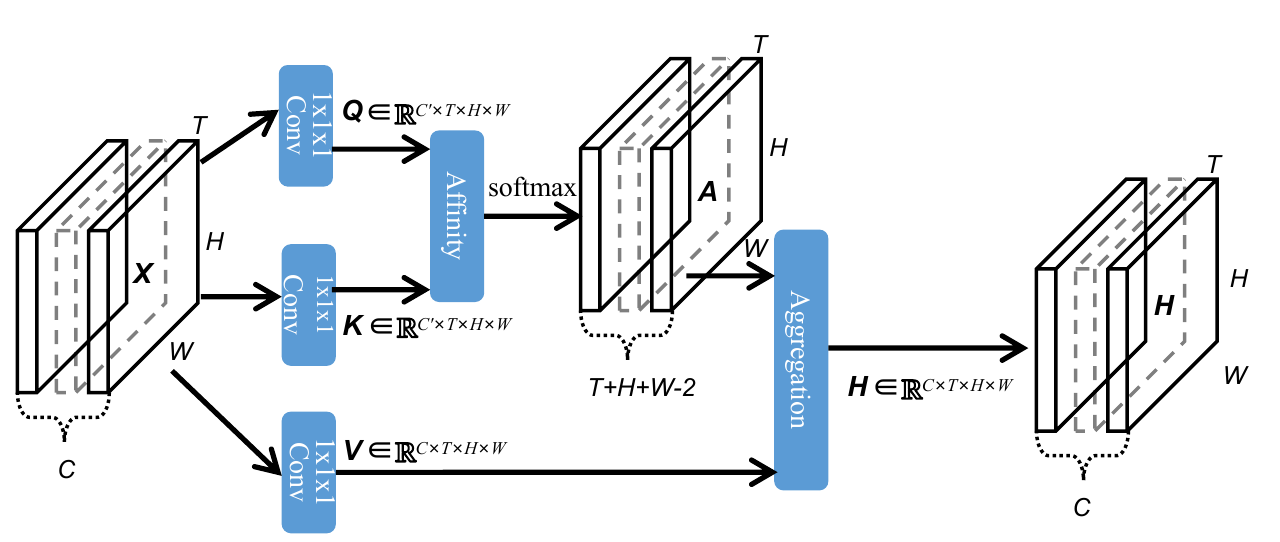}
	\caption{Details of our CCA-3D module. The 3D input feature map \textbf{\textsf{\textit{X}}} generates three feature maps \textbf{\textsf{\textit{Q}}}, \textbf{\textsf{\textit{K}}}, and \textbf{\textsf{\textit{V}}} through three different convolutional layers with $ 1\times1\times1 $ convolutional kernel size. \textbf{\textsf{\textit{Q}}} and \textbf{\textsf{\textit{K}}} generate the feature map \textbf{\textsf{\textit{A}}} via affinity operation. The feature map \textbf{\textsf{\textit{H}}} that aggregates long-range contextual information is obtained through the aggregation operation based on \textbf{\textsf{\textit{A}}} and \textbf{\textsf{\textit{V}}}.
	}
	\label{fig3}
\end{figure*}

\subsection{Criss-Cross Attention 3D Module}

A CCA-3D module is designed to collect contextual information in horizontal, vertical and depth directions. Its structure is shown in Figure \ref{fig3}. The 3D input feature map \textbf{\textit{X}} $\in \mathbb{R}^{C\times T\times H\times W}$ passes through two different 3D convolutional layers with kernels in the size of $ 1 \times 1 \times 1 $ to generate feature maps \textbf{\textit{Q}} and \textbf{\textit{K}} respectively, where \{\textbf{\textit{Q}}, \textbf{\textit{K}}\} $\in \mathbb{R}^{C'\times T\times H\times W}$. Here $C'$, which is smaller than $C$ for dimension reduction, represents the number of channels in the feature maps {\textbf{\textit{Q}} and \textbf{\textit{K}}}. Then \textbf{\textit{Q}} and \textbf{\textit{K}} generate attention map \textbf{\textit{A}} via \textbf{Affinity} \cite{Huang_2019_ICCV} and softmax operation which will be introduced in detail in the following paragraph. \textbf{\textit{X}} is fed into another convolutional layer with kernels in the size of $ 1 \times 1 \times 1$ to generate feature map \textbf{\textit{V}}$\in \mathbb{R}^{C\times T\times H\times W}$. We can get a new feature map \textbf{H} from feature map \textbf{\textit{A}} and \textbf{\textit{V}} by \textbf{Aggregation} operation, which will be also introduced in the following. Each point in \textbf{\textit{H}} aggregates long-range contextual information of the point with the other points on the same row, column and depth in the spatiotemporal domain.

\subsubsection{Affinity}
After \textbf{\textit{Q}} and \textbf{\textit{K}} are obtained, we can generate the feature map \textbf{\textit{A}}$\in \mathbb{R}^{(T+H+W-2) \times T\times H\times W}$ via affinity operation. For any point $u$ in the spatiotemporal dimension of feature map \textbf{\textit{Q}}, we can get a vector \textbf{\textit{Q}}$\mathbf{_u} \in \mathbb{R}^{C'}$. At the same time, we can also obtain a vector set $\mathbf{\Omega_u}\in \mathbb{R}^{(T+H+W-2) \times C'}$ containing a total of $(T + H + W - 2)$ elements from feature map \textbf{\textit{K}} by extracting feature vectors which are in the same row, column or depth of $u$. We denote the $i$th element of $\mathbf{\bm{\mathit{\Omega}}_u}$ as $\mathbf{\bm{\mathit{\Omega}}_{i,u}}\in \mathbb{R}^{C'}$, where $i \in [1, 2,...,T+H+W-2]$. Affinity operation is defined as follows:
\begin{equation}
	d_{i,u} = \mathbf{\bm{\mathit{Q}}_u} \mathbf{\bm{\mathit{\Omega}}_{i,u}^T},
\end{equation}
in which $d_{i,u} \in \mathbf{\bm{\mathit{D}}}$ is the degree of correlation between $\mathbf{\bm{\mathit{Q}}_u}$ and $\mathbf{\bm{\mathit{\Omega}}_{i,u}}$, and $\mathbf{\bm{\mathit{D}}} \in \mathbb{R}^{(T+H+W-2)\times T\times H\times W}$. Then, a softmax layer is applied on $\mathbf{\bm{\mathit{D}}}$ along the channel dimension to get the attention map $\mathbf{\bm{\mathit{A}}} \in \mathbb{R}^{(T+H+W-2)\times T\times H\times W}$. The $i$th element of the vector generated at position $u$ of the attention map \textbf{\textit{A}} is $\mathbf{\bm{\mathit{A}}}_{i,u}$:
\begin{equation}
	\bm{\mathit{A}}_{i,u} = \frac{e^{d_{i,u}}}{\sum\limits_{j=1}^{T+H+W-2}e^{d_{j,u}}}.
\end{equation}

\subsubsection{Aggregation}
For an arbitrary location $u$, we can also get the feature vector $\mathbf{\bm{\mathit{V}}_u} \in \mathbb{R}^{C}$ with all the points along the channels, and thus obtain a vector set $\mathbf{\bm{\mathit{\Phi}}_u} \in \mathbb{R}^{(T + H + W - 2)\times C}$. The vector set $\mathbf{\bm{\mathit{\Phi}}_u}$ is a collection of feature vectors in \textbf{\textit{V}} which are in the same row, column or depth of $u$. Long-range contextual information can be collected through aggregation operation:
\begin{equation}
	\bm{\mathit{H}}_u = \sum_{i=1}^{(T+H+W-2)}{\bm{\mathit{A}}_{i,u}\bm{\mathit{\Phi}}_{i,u}}
\end{equation}
where $\mathbf{\bm{\mathit{\Phi}}_{i,u}} \in \mathbb{R}^{C}$ is the $i$th vector in the vector $\mathbf{\bm{\mathit{\Phi}}}$ at position $u$. $\mathbf{\bm{\mathit{H}}_u} \in \mathbb{R}^{C}$ denotes a feature vector of the output feature map $\mathbf{\bm{\mathit{H}}}\in \mathbb{R}^{C\times T\times H\times W}$ at position $u$.

For any point in a 3D spatiotemporal feature map, a 3D criss-cross attention module collects the contextual information of all the points in horizontal, vertical and depth directions of the space where the point is located, \textit{i.e.} the points with the same coordinates in one dimension. Therefore, The proposed 3D criss-cross module only needs a small number of parameters, and the calculation amount is much less than that of non-local operation, resulting in little GPU memory usage.

\subsection{Recurrent Criss-Cross Attention 3D Module}

\begin{figure}[tp]
	\centering
	\subfigure[the first CCA-3D module]{
		\begin{minipage}[t]{0.42\textwidth} 
			\centering
			\includegraphics[width=0.55\textwidth]{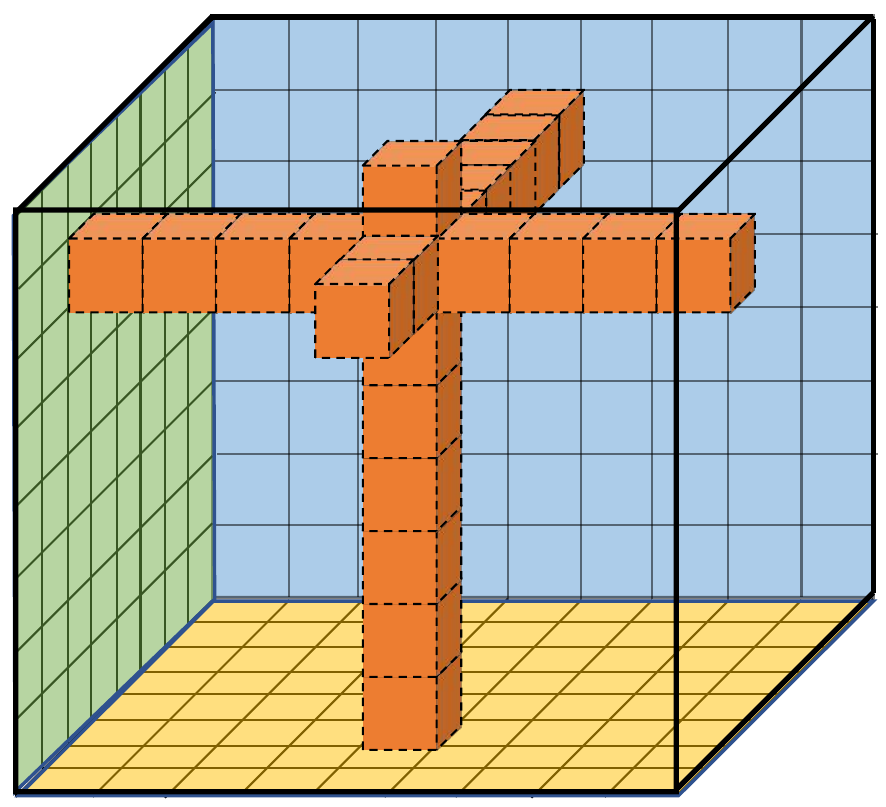}
			\label{fig4}
		\end{minipage}
	}
	\subfigure[the second CCA-3D module]{
		\begin{minipage}[t]{0.42\textwidth}
			\centering
			\includegraphics[width=0.55\textwidth]{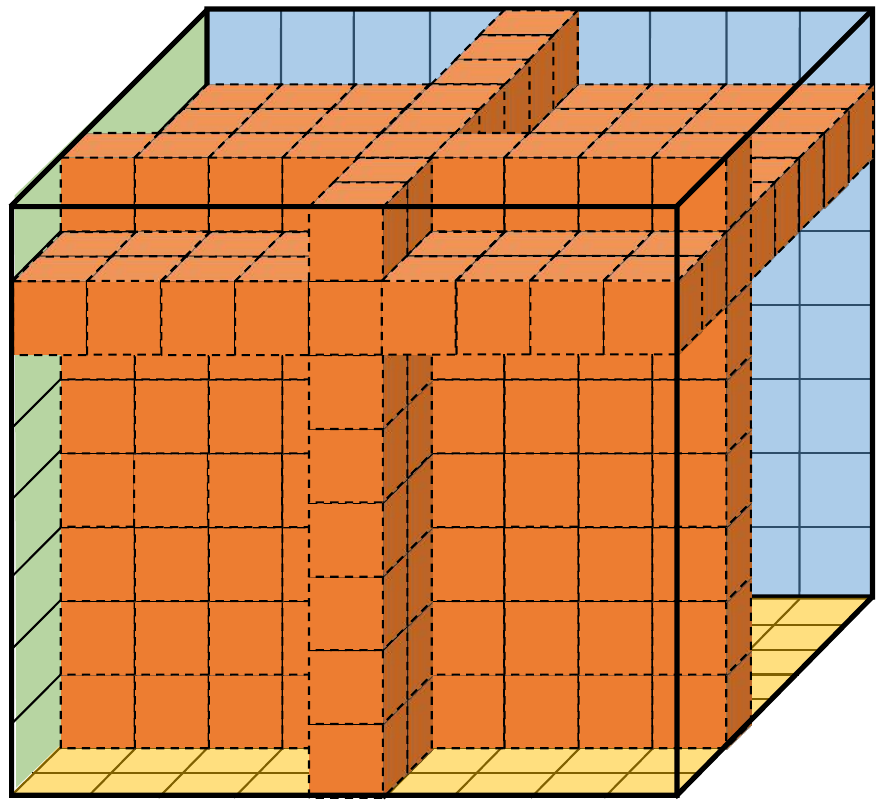} 
			\label{fig5}
		\end{minipage}
	}
	\centering
	\caption{The information propagation of a point in the first and second CCA-3D modules.}
\end{figure}

The information propagation of a point in a CCA-3D module is shown in Figure \ref{fig4}. It can be seen that the relationship between points captured by one CCA-3D module is sparse.

In order to get dense contextual information, we concatenate several CCA-3D modules to form the recurrent criss-cross attention 3D module, as shown in Figure \ref{fig2}. In an RCCA module, the first CCA-3D module takes the spatiotemporal feature map \textbf{\textit{X}} output from a CNN sub-network as input and outputs the feature map \textbf{\textit{H}} of the same shape. Each point in \textbf{\textit{H}} aggregates the relationship between the point and other points in the same row, column or depth. Considering that the appearance information of the input feature maps will be lost after each CCA-3D module harvesting the relationship between points, we add the input feature map \textbf{\textit{X}} to the output of each CCA-3D module, so that the succeeding layer is able to obtain both the relation and the appearance information. To balance the two information, \textbf{\textit{H}} is multiplied by a learnable weight scalar $\gamma$ and then added by \textbf{\textit{X}} to obtain the output \textbf{\textit{Y}}, \textit{i.e.},

\begin{equation}
	\mathbf{\bm{\mathit{Y}}}=\gamma\mathbf{\bm{\mathit{H}}}+\mathbf{\bm{\mathit{X}}}.
\end{equation}

\textbf{\textit{Y}} is fed into the second CCA-3D module, outputting a feature map \textbf{\textit{H'}} with the same shape as \textbf{\textit{H}}. Each point in \textbf{\textit{H'}} captures the information of all the points on the same plane in the spatial or temporal dimension, as shown in Figure \ref{fig5}. \textit{\textbf{H'}} is also multiplied by $\gamma$ and added by \textit{\textbf{X}} to get \textbf{\textit{Y'}}. \textbf{\textit{Y'}} is sent into the third CCA-3D module and generates \textbf{\textit{H''}}. Every point in \textbf{\textit{H''}} aggregates the relationship information between any two points in the spatiotemporal feature map. Therefore, in the spatiotemporal context modeling task, it is enough for an RCCA-3D module to contain three CCA-3D modules to harvest long-range dependencies and generate new feature maps with dense and rich contextual information. Finally, the output feature map of RCCA-3D module \textit{\textbf{Y''}} is obtained from \textbf{\textit{H''}} and fed into the next CNN sub-network. In an RCCA-3D module, all CCA-3D modules share weights so that they can make use of the learned relationship information, as well as reducing considerable extra parameters. Since the output of the RCCA-3D module has the same shape as the input, the proposed RCCA-3D module can be inserted into any 3D CNN architecture at any stage.

\subsection{Different Structures of RCCA-3D Module}

\begin{figure}[t]
	\centering
	\subfigure[structure \textit{b}]{
		\begin{minipage}{0.5\columnwidth}
			\centering
			\includegraphics[width=\columnwidth]{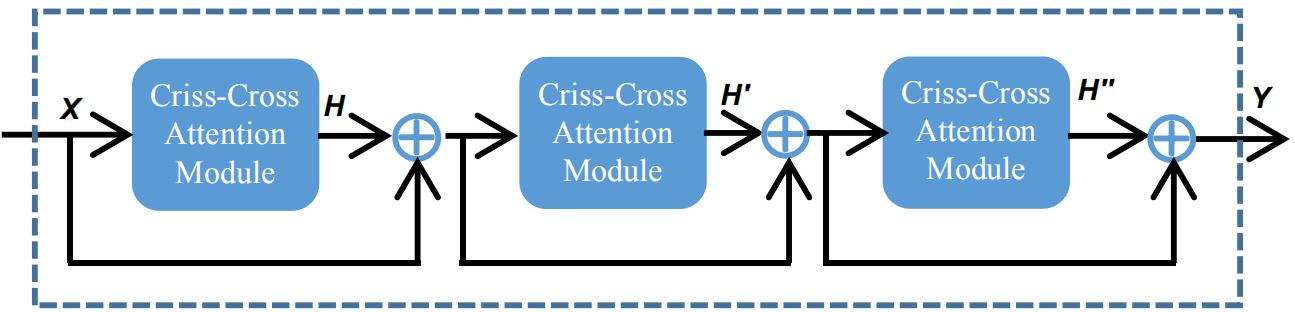}
			\label{fig_archb}
		\end{minipage}
	}
	\subfigure[structure \textit{c}]{
		\begin{minipage}{0.5\columnwidth}
			\centering
			\includegraphics[width=\columnwidth]{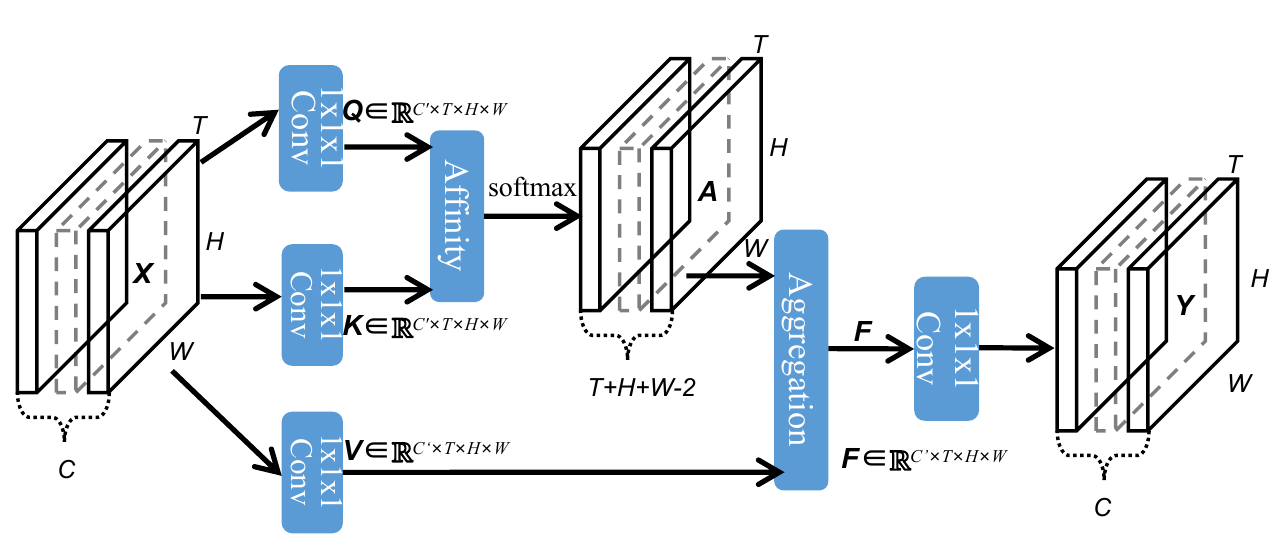}
			\label{fig_archc}
		\end{minipage}
	}
	\subfigure[structure \textit{d}]{
		\begin{minipage}{0.5\columnwidth}
			\centering
			\includegraphics[width=\columnwidth]{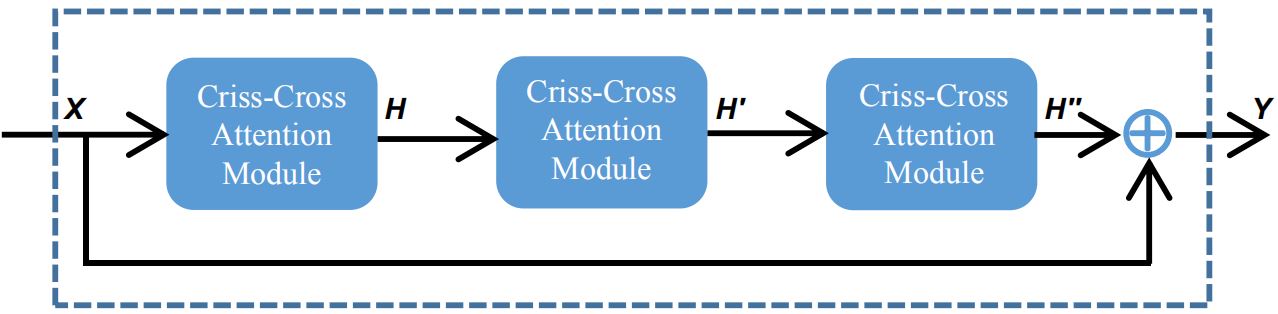}
			\label{fig_archd}
		\end{minipage}
	}
	\centering
	\caption{Illustration of the RCCA-3D structures \textit{b}, \textit{c} and \textit{d}.}
	\label{fig_archs}
\end{figure}

In order to research the fusion mechanism of relation and appearance, we design four different structures of the RCCA-3D module. We define the RCCA-3D structure in Figure \ref{fig2} as structure \textit{a}, which can be denoted as:
\begin{equation}
	\mathbf{\bm{\mathit{Y''}}}= \gamma \mathcal{F}(\gamma \mathcal{F}(\gamma \mathcal{F}(\mathbf{\bm{\mathit{X}}})+\mathbf{\bm{\mathit{X}}})+\mathbf{\bm{\mathit{X}}})+\mathbf{\bm{\mathit{X}}}
\end{equation}
in which $\mathcal{F}()$ represents the operations of the CCA-3D module.
The other three structures are named as structure \textit{b}, \textit{c} and \textit{d}, as shown in Figure \ref{fig_archs}.

Structure \textit{b} is defined as:
\begin{equation}
	\begin{cases}
		\mathbf{\bm{\mathit{Y''}}}=\gamma \mathcal{F}(\mathbf{\bm{\mathit{Y'}}}) + \mathbf{\bm{\mathit{Y'}}} \\
		\mathbf{\bm{\mathit{Y'}}}=\gamma \mathcal{F}(\mathbf{\bm{\mathit{Y}}}) + \mathbf{\bm{\mathit{Y}}} \\
		\mathbf{\bm{\mathit{Y}}}=\gamma \mathcal{F}(\mathbf{\bm{\mathit{X}}}) + \mathbf{\bm{\mathit{X}}}
	\end{cases}
\end{equation}
where the input of each CCA-3D module is added to its output to form a residual connection. It is the same as the 2D structure in \cite{Huang_2019_ICCV}.
It can be seen as  a direct extension of \cite{Huang_2019_ICCV}.   
Our experiments show that the performance of the structure \textit{b} for action recognition task is worse than our proposed structure \textit{a}.

The difference between structure \textit{c} and structure \textit{a} lies in the CCA-3D module. In the CCA-3D module of structure \textit{c}, dimension reduction is also applied on the convolutional layer of \textbf{\textit{V}}, so \textbf{\textit{V}} has $ C' $ channels. Generally, $ C' $ is smaller than $ C $, causing channel reduction of the feature map output by aggregation operation. Therefore, after the aggregation operation, a convolutional layer with kernel size of $ 1 \times 1 \times 1 $ is added to restore the original channel number $ C $.

Structure \textit{d} is defined as:
\begin{equation}
	\mathbf{\bm{\mathit{Y''}}}=\gamma\mathcal{F}(\gamma\mathcal{F}(\gamma\mathcal{F}(\mathbf{\bm{\mathit{X}}})))+\mathbf{\bm{\mathit{X}}}
\end{equation}
where only the first CCA-3D module is input with the original appearance information, but \textit{\textbf{Y''}} obtains the original appearance information by adding \textit{\textbf{X}} before output. The performance of these four structures are compared in the next section.

\section{Experiments}
We adopt TSM \cite{lin2019tsm} and MF-Net \cite{Chen2018ECCV} as our backbones, and use the popular UCF101 \cite{soomro2012ucf101}, HMDB51 \cite{kuehne2011hmdb}, Mini-Kinetics and Kinetics-400 \cite{kay2017kinetics} datasets as evaluation benchmarks.
The experiments show that the RCCA-3D module achieves higher accuracy in video-based action recognition tasks with lower memory usage and less parameters compared with the non-local module. We explore the internal structure of the RCCA module and the inserted location in networks to maximize RCCA's ability of spatiotemporal context modeling.

\subsection{Datasets and Network Structure}

UCF101 \cite{soomro2012ucf101} is an action recognition dataset of realistic action videos, collected from YouTube. It contains 13320 videos that belong to 101 categories. HMDB51 \cite{kuehne2011hmdb} dataset contains 6849 videos divided into 51 action categories, each containing a minimum of 101 videos. Most of the videos are extracted from movies, and a small proportion from public databases. These two datasets both have three training/testing splits. We use the first split for ablation studies and all the three splits for comparisons with other methods.
Kinetics-400 \cite{kay2017kinetics} involves 400 human action categories and contains about 246k training videos and 20k validation videos. All the videos are from YouTube. Following \cite{lin2019tsm} and \cite{wang2018non}, we use the training set in training stage and the validation set in testing stage. Mini-Kinetics \cite{Carreira2017CVPR} is a subset of Kinetics-400 and contains 200 classes of videos selected from Kinetics-400. Its data size is about 6 times that of the UCF101 dataset but still much smaller than that of the Kinetics-400.

TSM is based on ResNet-50 structure while MF-Net is based on ResNet-34 structure. The structures of the two networks both contain 5 convolution stages that are composed of 1, 3, 4, 6 and 3 residual blocks respectively. Generally, these 5 convolution stages are named in order according to conv\textit{1} - conv\textit{5}.
If a module is inserted into the \textit{i}th residual block of the \textit{n}th stage, then the inserted position of this module is represented as conv\textit{n}\_\textit{i}.
For example, if an RCCA-3D module is inserted into the fifth residual block in the conv\textit{4} stage, the inserted location is conv\textit{4}\_\textit{5}.
According to the extension experiment of TSM \cite{lin2019tsm} with non-local mechanism, 5 positions are used to insert non-local modules in total, which are conv\textit{3}\_\textit{1}, conv\textit{3}\_\textit{3}, conv\textit{4}\_\textit{1}, conv\textit{4}\_\textit{3}, and conv\textit{4}\_\textit{5}. We verify that our RCCA-3D achieves good performance when inserted into different positions. So as to make a fair comparison with other methods, we also retain those five locations according to the settings of TSM.
Besides, we try more locations in other stages to study the performance of our method.

\subsection{Training and Testing Settings}

By default, all experiments are conducted on two NVIDIA GeForce RTX 2080Ti GPUs with limited training batch size, and we adopt our own implementation as the baselines. 
Given that the batch size has a significant impact on accuracy, we accumulate the gradient of several batches of data and reduce the learning rate to make the small batches equivalent to a larger batch.
For example, if the maximum batch size is 16, and our goal is to expand the batch size to 128, we reduce the learning rate to 1/8 of the original and accumulate the gradient calculated by 8 batches of the training data. Finally we update the parameters of the whole network once for every 8 batches.
These operations are implemented using PyTorch \cite{paszke2019pytorch}.

We keep to use the same training settings with the original backbone networks plugged in RCCA-3D modules.
For TSM + RCCA-3D experiments, 8 frames are sampled as a clip by uniform sampling and the shift proportion of the temporal shift module is 1/8. Based on the training settings of \cite{lin2019tsm}, dense sampling can achieve higher accuracy, but its computation cost is 10 times higher than that of the uniform sampling. To reduce the training cost, we use uniform sampling and make comparisons with the results of uniform sampling in \cite{lin2019tsm}.
For the Kinetics dataset, there are two testing conditions according to the open source code of \cite{lin2019tsm}. In the training stage, 1 crop is always sampled for each clip. In the testing stage, 1 crop and 10 crops are sampled per clip. We report the results that sampling 10 crops in the testing stage by default.
We set the initial learning rate as 0.02. At the 40th and 80th training epoch, the learning rate is reduced with a factor of 0.1. The weight decay is set to 5e-4 and we train the network for 100 epochs in total.
Regarding MF-Net + RCCA-3D experiments, we sample 16 frames as a clip in a dense manner. The initial learning rate is set to 0.05, which is reduce at the 60th, 90th and 110th epoch with a factor of 0.1. The weight decay is set to 1e-4 and the total training process takes 120 epochs.

For each experiment, all frames are randomly scaled, randomly flipped and resized to $224 \times 224$ before fed into the network. Due to the limitation of computing capacity, and to be consistent with the settings of the other works, we do not use the optical flow modality.
All the networks are optimized by SGD with a momentum of 0.9. The batch size is set to 16 for each training step, but it is equivalent to a batch size of 128 by gradient accumulation.
The learnable weight $\gamma$ is initialized as 1. We use the backbone models pre-trained on ImageNet-1k \cite{krizhevsky2012imagenet} to initialize the parameters of the networks except the RCCA-3D modules. We report the top-1 accuracy of all experiments in video level, among which the video-level label is obtained by averaging the clip predictions.

\subsection{Ablation Studies}

\begin{table}[tp]
	\centering
	\caption{Results of different structures of RCCA-3D module on UCF101 dataset.}
	\renewcommand\tabcolsep{4.0pt}
	\begin{tabular}{@{}lcrrr@{}}
		\toprule
		Model      	& Structure 			& FLOPs  			& Param 		& Top-1    \\ \midrule
		TSM        	& -         			& 33.0G / 100.0\%   & 24.30M / 100.0\%      & 85.01\% \\
		TSM + RCCA	& \textit{a}   			& 49.3G / 149.4\% 	& 24.69M / 101.6\%   	& 86.81\% \\
					& \textit{b}       	  	& 49.3G / 149.4\% 	& 24.69M / 101.6\%   	& 85.46\% \\
					& \textit{c}        	& 49.0G / 148.5\% 	& 24.69M / 101.6\%    	& 85.30\% \\
					& \textit{d}         	& 49.3G / 149.4\% 	& 24.69M / 101.6\%   	& 84.99\% \\ \bottomrule
	\end{tabular}
	\label{table1}
\end{table}

We use TSM as the baseline network in our ablation studies. The amount of calculation is measured by FLOPs, \textit{i.e.} floating point multiplication adds, and the amount of parameters is measured by the number of parameters in training. Absolute FLOPs and parameters are reported in the experiments with the proposed RCCA-3D module.
We also report the percentage to make an intuitive comparison with the baseline.

\textbf{Structures:} We insert one RCCA-3D module of the four different structures \textit{a}, \textit{b}, \textit{c}, and \textit{d} into the baseline network at conv\textit{3}\_\textit{3}, whose results can be seen in Table \ref{table1}. We set the number of inner channels in each CCA-3D module as a quarter of that in the input feature map. It can be seen that the increased parameter amount of the four structures are negligible. TSM has 33G FLOPs and 24.3M parameters \cite{lin2019tsm}. Introducing one RCCA-3D module brings about 16G (49\%) extra FLOPs and 0.39M (1.6\%) parameters. Compared with structure \textit{b} and structure \textit{d}, structure \textit{a} makes full use of the input appearance information, indicating that the original appearance feature information plays an important role in capturing long-range dependencies using RCCA-3D module. Although structure \textit{c} can reduce considerable FLOPs if $ C' $ is set smaller, it cannot make an ideal accuracy gain. Since the accuracy of structure \textit{a} is much higher than that of the others, we adopt structure \textit{a} as the standard structure of the RCCA-3D module.

\begin{table}[tp]
	\centering
	\caption{Comparisons of different inserted stages of RCCA-3D.}
	\renewcommand\tabcolsep{4.0pt}
	\begin{tabular}{@{}lcrrr@{}}
		\toprule
		Model      	& Position 						& FLOPs 				& Param 				& Top-1    \\ \midrule
		TSM        	& -        						& 33.0G / 100.0\%    	& 24.30M / 100.0\%    	& 85.01\% \\
		TSM + RCCA 	& conv\textit{3}\_\textit{3}  	& 49.3G / 149.3\%		& 24.69M / 101.6\%   	& 86.81\% \\
					& conv\textit{4}\_\textit{5}  	& 48.2G / 146.1\%		& 25.87M / 106.5\%    	& 86.15\% \\
					& conv\textit{5}\_\textit{2}  	& 47.9G / 145.2\%		& 30.59M / 125.9\%   	& 84.85\% \\ \bottomrule
	\end{tabular}
	
	\label{table4}
\end{table}

\begin{table}[tp]
	\centering
	\caption{Comparisons of different numbers of CCA-3D modules (\textit{R}) in an RCCA-3D module on UCF101 dataset.}
	\begin{tabular}{@{}lcrrr@{}}
		\toprule
		Model      		& \textit{R} & FLOPs 		& Param 				& Top-1    \\ \midrule
		TSM        		& - 	& 33.0G / 100.0\%   & 24.30M / 100.0\%    	& 85.01\% \\
		TSM + RCCA 		& 1 	& 38.4G / 116.4\% 	& 24.69M / 101.6\%		& 84.91\% \\
						& 2 	& 43.9G / 133.0\%	& 24.69M / 101.6\%      & 85.73\% \\
						& 3 	& 49.3G / 149.3\%	& 24.69M / 101.6\%      & 86.81\% \\
						& 4 	& 54.7G / 165.8\%	& 24.69M / 101.6\%      & 85.78\% \\ \bottomrule
	\end{tabular}
	\label{table2}
\end{table}

\begin{table}[tp]
	\centering
	\caption{Comparisons of different channel reduction fractions in RCCA-3D at conv\textit{3} stage.}
	\begin{tabular}{@{}lcrrr@{}}
		\toprule
		Model      	& \textit{C\_d} & FLOPs 			& Param 	   		   & Top-1    \\ \midrule
		TSM        	& -    			& 33.0G / 100.0\%   & 24.30M / 100.0\%     & 85.01\% \\
		TSM + RCCA 	& 1/2  			& 54.5G / 165.2\% 	& 24.83M / 102.2\%     & 84.30\% \\
					& 1/4  			& 49.3G / 149.4\%	& 24.69M / 101.6\%     & 86.81\% \\
					& 1/8  			& 46.7G / 141.5\%	& 24.63M / 101.4\%     & 85.22\% \\
					& 1/16 			& 45.4G / 137.6\%	& 24.60M / 101.2\%     & 85.73\% \\ \bottomrule
	\end{tabular}
	
	\label{table5}
\end{table}

\textbf{Inserted stages:} To explore the stage most suitable for RCCA-3D to plug into, we insert one RCCA-3D module into different stages of TSM respectively. Following non-local neural network \cite{wang2018non}, each module is placed before the last residual unit in each stage. The results are shown in Table \ref{table4}. We find that the RCCA-3D module has the highest accuracy when inserted into conv\textit{3}. Although the calculation amount of RCCA-3D in conv\textit{3} is slightly more than that in other stages, the parameter amount is much less. Experiment of inserting RCCA-3D module into stage 2 is not conducted because it leads to too much computation cost (about 53.5G FLOPs). Moreover, the feature maps in conv\textit{2} stage have not enough abstract semantic representations \cite{NIPS20187318}, so it is difficult to extract semantic contextual information and does few favor to spatiotemporal context modeling.

\textbf{Number of modules:} We also experiment on the number of CCA-3D modules, which is denoted as \textit{R}, in the RCCA-3D module at conv\textit{3}\_\textit{3}. The results are shown in Table \ref{table2}. With the increase of the number of CCA-3D modules, the extra amount of calculation is increased linearly.
Specifically, if \textit{R} is increased by 1, the extra FLOPs ($\triangle$FLOPs) is increased by 5.4G.
Meanwhile, the prediction accuracy increases gradually when \textit{R} is not more than 3. When \textit{R} is equal to 4, the $\triangle$FLOPs has a maximum of 21.7G but the accuracy decreases compared with the accuracy when \textit{R} is equal to 3. When \textit{R} is 3, the spatiotemporal modeling ability of RCCA-3D reaches the optimal level and the network achieves the highest accuracy 86.81\% with 16.3G $\triangle$FLOPs, which is consistent with our theoretical analysis.

\textbf{Inner channels:} Channel reduction is an effective and direct way to reduce the amount of calculation. Therefore, we insert an RCCA-3D module with channel reduction into conv\textit{3}\_\textit{3} to find the number of inner channels that leads to higher accuracy and less amount of computation. The results are shown in Table \ref{table5}. \textit{C\_d} is a fraction representing that the number of channels in the CCA-3D module is \textit{C\_d} times that in the input feature map, \textit{i.e.} $ C' $/ $ C $ = \textit{C\_d} in Figure \ref{fig3}. From Table \ref{table5}, it can be seen that as the number of channels decreases exponentially, the amount of calculation and parameters decreases more and more slowly because the number of input and output channels of the convolutional layer \textit{\textbf{V}} in the CCA module is always the same. The highest accuracy is achieved when \textit{C\_d} = 1/4, so we adopt \textit{C\_d} = 1/4 in other experiments.

\subsection{Comparisons with Non-local Network}

\begin{table}[tp]
	\centering
	\caption{Comparisons on different datasets and different networks between RCCA-3D modules and non-local modules.}
	\renewcommand\tabcolsep{2.0pt}
	\begin{tabular}{@{}lcccc@{}}
		\toprule
		\multirow{2}{*}{Method} 	  & \multirow{2}{*}{FLOPs} & \multirow{2}{*}{Param} & \multicolumn{2}{c}{Top-1} \\ \cmidrule(l){4-5}
							&						  &						   & UCF101		 & HMDB51     \\ \midrule
		TSM               	& 33.0G / 100.0\%         & 24.30M / 100.0\%       & 85.01\%     & 50.07\%    \\
		TSM+NL*1            & 56.4G / 170.9\%         & 24.83M / 102.2\%       & 86.75\%     & 52.62\%    \\
		TSM+RCCA*1          & 49.3G / 149.4\%         & 24.69M / 101.6\%       & 86.81\%     & 52.55\%    \\ \midrule
		MF-Net             	& 11.1G / 100.0\%         & ~8.00M / 100.0\%       & 78.60\%     & 39.56\%    \\
		MF-Net+NL*5         & 24.7G / 222.5\%         & ~9.04M / 113.0\%       & 80.43\%     & 41.17\%    \\
		MF-Net+RCCA*5       & 23.1G / 208.1\%         & ~8.78M / 109.8\%       & 80.76\%     & 41.52\%    \\ \bottomrule
	\end{tabular}
	\label{table6}
\end{table}

We conduct sufficient experiments to make a comprehensive comparison between the RCCA-3D module and the non-local module. Non-local modules and RCCA-3D modules are inserted into TSM and MF-Net respectively, whose results are shown in Table \ref{table6}.

As to TSM, we select three locations in different stages, \textit{i.e.}, conv\textit{3}\_\textit{3}, conv\textit{4}\_\textit{5} and conv\textit{5}\_\textit{2}, to insert one module at a time. We pick out the models with the highest accuracies to report, whose results are obtained by inserting in conv\textit{3}\_\textit{3}. As shown in Table \ref{table6}, the accuracies of the network with one RCCA-3D module and the network with one non-local module are roughly the same on the UCF101 and HMDB51 datasets. However, the FLOPs of the network with our RCCA-3D module is 30\% less than that of the network with the non-local module, and the number of parameters is 25\% less. Since MF-Net is more lightweight than TSM, we insert 5 non-local modules or RCCA-3D modules in conv\textit{3}\_\textit{1}, conv\textit{3}\_\textit{3}, conv\textit{4}\_\textit{1}, conv\textit{4}\_\textit{3} and conv\textit{4}\_\textit{5} for comparison. It should be noted that, we use the downsampling trick for the non-local modules, resulting in lots of FLOPs reduction. However, MF-Net equipped with RCCA-3D modules still costs less calculation and parameters than that of non-local modules, and achieves higher accuracy. The results in Table \ref{table6} confirms the superiority of the proposed RCCA-3D module.

\begin{table}[tp]
	\centering\renewcommand\tabcolsep{5.0pt}
	\caption{Results of TSM with non-local and RCCA modules on Mini-Kinetics, UCF101 and HMDB51 datasets. The networks are pre-trained on Mini-Kinetics.}
	\begin{tabular}{@{}lccc@{}}
		\toprule
		\multirow{3}{*}{Method}                 		& \multicolumn{3}{c}{Top-1} 	\\ \cmidrule(l){2-4}
								& Mini-Kinetics & UCF101      	& HMDB51     				\\ \midrule
		TSM                 	& 75.02\% 		& 91.38\%  		& 62.48\%    	\\
		TSM + NL            	& 76.11\%		& 91.62\%     	& 63.44\%    	\\
		TSM + RCCA-3D           & 76.38\% 		& 92.52\%      	& 63.53\%    	\\  \bottomrule
	\end{tabular}
	\label{table8}
\end{table}

\begin{table}[tp]
	\caption{Comparisons of TSM and TSM with RCCA-3D module on the Kinetics-400 dataset.}
	\label{table_AA}
	\centering
	\begin{tabular}{@{}lcc@{}}
		\toprule
		\multirow{2}{*}{Model} & \multicolumn{2}{c}{Top-1} \\ \cmidrule(l){2-3}
							   & 1-crop            & 10-crop          \\ \midrule
		TSM                    & \textit{71.2\%} \cite{lin2019tsm}     &\textit{72.8\%} \cite{lin2019tsm}    \\
		TSM + RCCA-3D          & 71.6\%            & 73.6\%           \\ \bottomrule
	\end{tabular}
\end{table}

\begin{table*}[bp]
	\caption{Results of fine-tuning on UCF101 and HMDB51 datasets. S\# denotes the split setting of the dataset. Avg denotes the average precision of all the splits.}
	\label{table_BB}
	\centering
	\begin{tabular}{@{}llcccccccc@{}}
		\toprule
		\multirow{2}{*}{Model} 	& \multirow{2}{*}{Pre-training} 	& \multicolumn{4}{c}{UCF101 (Top-1)} & \multicolumn{4}{c}{HMDB51 (Top-1)} \\ \cmidrule(l){3-10}
		&									& S1      & S2     & S3     & Avg    & S1      & S2     & S3     & Avg    \\ \midrule
		TSM                    	& Kinetics-400                  	& 94.6\%  & -      & -      & -      & 70.6\%  & -      & -      & -      \\
		TSM + RCCA-3D          	& Kinetics-400                  	& 95.9\%  & 95.0\% & 94.4\% & 95.1\% & 70.9\%  & 69.7\% & 70.0\% & 70.2\% \\
		TSM + RCCA-3D          	& ImageNet                      	& 86.8\%  & 86.1\% & 86.0\% & 86.3\% & 52.6\%  & 52.9\% & 52.3\% & 52.6\% \\ \bottomrule
	\end{tabular}
\end{table*}

It should be noted that due to limitations of the hardware in this section, we only use the models pre-trained on ImageNet for initialization, so the baseline accuracies are much lower than that reported in \cite{Chen2018ECCV} and \cite{lin2019tsm} which are pre-trained on large-scale video dataset. After training the MF-Net backbone on Mini-Kinetics dataset and then fine-tuning on UCF101 and HMDB51 datasets, the baseline accuracy on the two datasets can increase from 78.6\% and 39.6\% to 91.5\% and 65.6\%, respectively. More experiments with pre-training on the Mini-Kinetics and Kinetics-400 datasets will be introduced in the next section.

\subsection{Pre-training on Video Dataset}
In order to verify that our method can also benefit from pre-training on large-scale video dataset, we conduct a series of experiments on TSM with pre-training on Mini-Kinetics
and Kinetics-400.

The results on Mini-Kinetics can be seen in Table \ref{table8}. The TSM with RCCA-3D module achieves the highest accuracy.
The results on the UCF101 and HMDB51 datasets with the Mini-Kinetics pre-training are also shown in Table \ref{table8}.
For the backbone, the accuracy is increased to 91.38\% (+6.37\%) on UCF101 and 62.48\% (+12.41\%) on HMDB51 with pre-training on the Mini-Kinetics.
Then We insert the RCCA-3D and non-local module into conv\textit{3}\_\textit{3} and conv\textit{4}\_\textit{5} respectively. With Mini-Kinetics pre-training, TSM with RCCA-3D achieves an accuracy of 92.52\% on UCF101 and 63.53\% on HMDB51. It is consistently higher than that of TSM with non-local module. 

The performances of the baseline and our model on the Kinetics-400 dataset are shown in Table \ref{table_AA}, where 4 NVIDIA GeForce 2080Ti GPUs are used for training.
It can be seen that the accuracy of the TSM has been improved by being equipped with the RCCA-3D module on Kinetics-400 dataset. Compared with the results of \cite{lin2019tsm}, our accuracy are improved by 0.4\% and 0.8\% respectively under the testing conditions of 1-crop and 10-crop. It should be noted that, limited by the hardware resources, the maximum batch size of our experiments is 32, which is half of the setting in \cite{lin2019tsm}.
If trained on more GPUs with a larger batch size, the performance of our method should be higher. It shows that our RCCA-3D module is robust for spatiotemporal modeling, and the performance of the model is improved.

Then we fine-tune our model on the UCF101 and HMDB51 datasets. Pre-training on large video dataset  provides a significant boost for the models. As shown in Table \ref{table_BB}, compared with the previous model without being pre-trained on the Kinetics-400 dataset, the pre-trained model improves the accuracies by 9.1\% and 18.3\% on the UCF101 and HMDB51 datasets respectively. With larger video dataset pre-training, the model achieves higher absolute accuracy for action recognition.

\subsection{Comparisons with the State-of-the-art}

\begin{table}[tp]
	\centering\renewcommand\tabcolsep{2.0pt}
	\caption{Comparisons with state-of-the-art results on the UCF101, HMDB51 and Kinetics datasets. In the first part, the results are the average accuracy on the three splits. In the second part, only the accuracy on the first split is reported. * indicates that the model is pre-trained on large-scale video datasets.}
	\begin{tabular}{@{}lrrr@{}}
		\toprule
		\multirow{2}{*}{Method}                 		& \multicolumn{3}{c}{Top-1} 				\\ \cmidrule(l){2-4}
		& UCF101      	& HMDB51		& Kinetics  \\ \midrule
		GRU-RCN \cite{ballas2015delving}               	& 80.7\%      	& -          	& - 		\\
		T-ResNet \cite{feichtenhofer2017temporal}       & 85.4\%      	& 48.9\%     	& - 		\\
		Res3D* \cite{tran2017convnet}             		& 85.8\%     	& 54.9\%     	& - 		\\
		TSN \cite{wang2018temporal}						& -				& -				& 69.9\%	\\
		TSM \cite{lin2019tsm}                        	& 84.8\%     	& 50.6\%    	& 72.8\% 	\\
		TSM + NL \cite{wang2018non}                   	& 85.6\%     	& 52.3\%    	& -		 	\\
		TSM + RCCA-3D (Our method)                      & 86.3\%      	& 52.6\%     	& 73.6\%	\\ \midrule
		TSN* \cite{wang2018temporal}					& 94.0\%		& {68.5\%}		& -			\\
		TSM* \cite{lin2019tsm}  						& 94.6\% 		& 70.6\%    	& - 		\\
		TSM + RCCA-3D* (Our method)  					& 95.9\%  		& 70.9\%     	& - 		\\  \bottomrule
	\end{tabular}
	\label{table7}
\end{table}

We compare our method with the others only using RGB modality (except Res3D and TSN) for a fair comparison. We calculate the average accuracy of all the splits on the UCF101, HMDB51 and Kinetics-400 datasets. The results are shown in the first part of Table \ref{table7}.
It should be noted that the Res3D* is pre-trained on Sports1M \cite{karpathy2014large} and the TSN uses both RGB and optical flow modalities. Although the Res3D* is pre-trained on large-scale video dataset, it has a lower accuracy than our method without large-scale video dataset pre-training on the UCF101 dataset.
We cannot satisfy the original experimental conditions of \cite{lin2019tsm}, which has 8 NVIDIA Tesla P100 GPUs. However, the accuracy of the model with our method is still improved by 0.8\% compared with the baseline. It will perform better if the same experimental setup is maintained.

The TSN*, TSM* and TSM + RCCA* are pre-trained on Kinetics-400 and fine-tuned on the first splits of UCF101 and HMDB51, whose results are shown in the second part of Table \ref{table7}. We only report the accuracy on the first split since \cite{lin2019tsm} only experiments on the first split. It can be seen our TSM + RCCA-3D model achieves the best performance.

\subsection{Visualization}

\begin{figure}[tp]
	\centering
	\subfigure[input feature maps]{
		\begin{minipage}[t]{\linewidth}
			\centering
			\includegraphics[width=8.5cm]{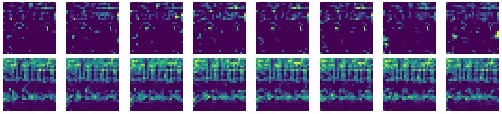}
		\end{minipage}
	}
	\subfigure[output feature maps]{
		\begin{minipage}[t]{\linewidth}
			\centering
			\includegraphics[width=8.5cm]{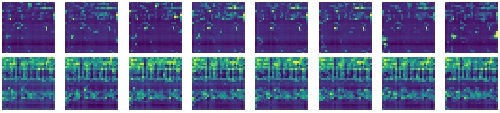}
		\end{minipage}
	}
	\centering
	\caption{Visualization of the input and output feature maps of the RCCA-3D module. Each row is composed of the feature maps arranged chronologically from left to right of a channel.}
	\label{fig_ccainout}
\end{figure}

\begin{figure*}[tp]
	\centering
	\subfigure[sweeping the floor]{
		\begin{minipage}[tp]{\textwidth}
			\centering
			\includegraphics[width=0.95\textwidth]{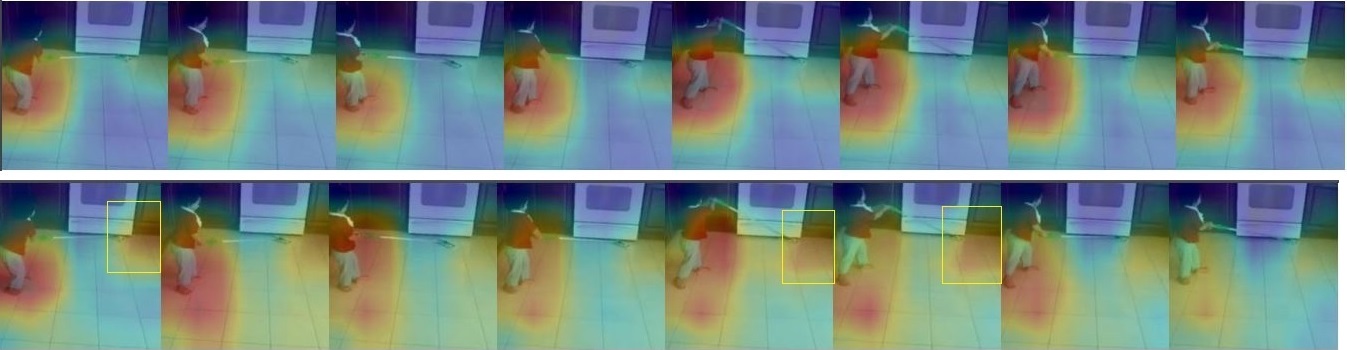}
		\end{minipage}
		\label{cama}
	}
	\subfigure[playing the violin]{
		\begin{minipage}[tp]{\textwidth}
			\centering
			\includegraphics[width=0.95\textwidth]{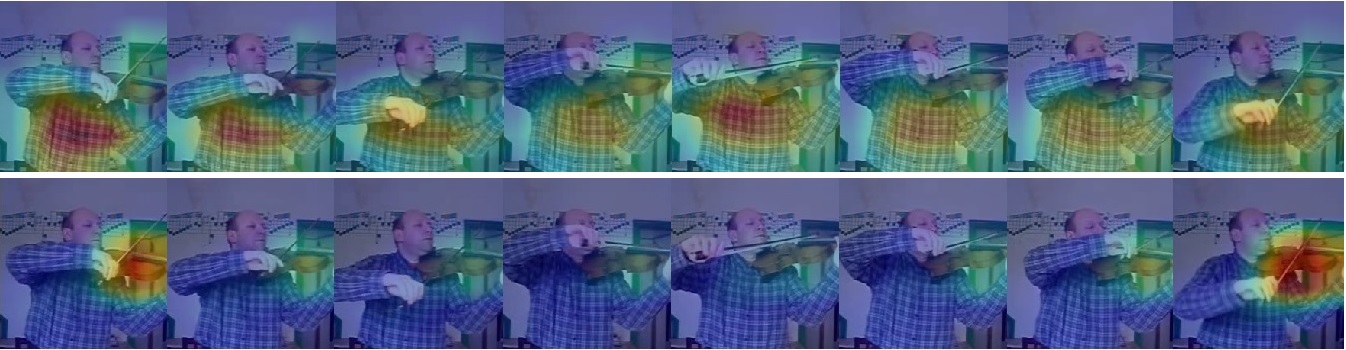}
		\end{minipage}
		\label{camb}
	}
	\centering
	\caption{Visualization of the the class activation maps for two clips of videos. We flatten each video chronologically into frames. The first row is generated by TSM and the second row is generated by TSM with RCCA-3D in each subfigure.}
	\label{figcam}
\end{figure*}

To understand the effects brought by RCCA-3D module, we display the input and output feature maps of the RCCA-3D module, as shown in Figure \ref{fig_ccainout}. The input is a clip of basketball playing video. We can see that more relevant points in the output feature maps are brighter than the input feature maps. In other words, the locations of those points are added on the long-range relationship information captured by the RCCA-3D module, so the next layer will get more semantic information, which is why RCCA-3D module works.

We also visualize the class activation maps \cite{zhou2016learning} (CAM) of the TSM and TSM+RCCA-3D models, which is shown in Figure \ref{figcam}. The original CAM is used for the visualization of the relevance between each pixel and the class in images. We extend that to visualizing the relevance of pixels and the class in videos, instead of applying the original CAM to each single frame. Figure \ref{cama} shows a boy sweeping the floor. We can see that the TSM almost only attends to the boy from the first row. The second row displays that the mop on the floor is also attended to obviously by TSM with RCCA-3D module, which is marked with a yellow box. Figure \ref{camb} shows a man playing the violin. In the first row, TSM pays most attention to the body of the man rather than his hand and the violin, while TSM with RCCA-3D in the second row mainly attends to the violin and the hand area. That is to say, adding RCCA-3D module makes the related points in spatiotemporal space more salient, and contributes to spatiotemporal context modeling for action recognition.

\section{Conclusion}

In this paper, we have presented an efficient long-range {context} learning model for action recognition. Instead of directly modeling the relationship between any two points in the spatiotemporal feature map, we extend 2D criss-cross attention (CCA) to 3D to model the pair-wise relation for the points in the same line, which is a 3D criss-cross path in spatiotemporal space, each time. By stacking the CCA-3D modules with recurrent connection, the relation of points is transmitted from line to plane, finally to the whole spatiotemporal space. Therefore, the dense global contextual information is reconstructed with the factorized sparse relation maps, which can reduce the computation and memory cost significantly. The proposed recurrent CCA-3D (RCCA-3D) module can be plugged into existing networks conveniently. We make extensive experiments with different action recognition networks as backbones to compare with other relation modeling methods and determine the optimal structure of the proposed module. Plugged with our RCCA-3D module, the performance is boosted while the cost is minor. Promising experimental results on multiple datasets {demonstrate} the efficiency and effectiveness of our model.

\bibliography{mybibfile}

\end{document}